\documentclass[10pt,twocolumn,letterpaper]{article}

\usepackage[pagenumbers]{cvpr} 
\usepackage{cuted}
\usepackage{tabularx}
\usepackage{multirow}
\usepackage[dvipsnames]{xcolor}

\definecolor{cvprblue}{rgb}{0.21,0.49,0.74}
\usepackage[pagebackref,breaklinks,colorlinks,citecolor=cvprblue]{hyperref}

\def\paperID{4720} 
\def\confName{CVPR}
\def\confYear{2024}

\title{Sherpa3D: Boosting High-Fidelity Text-to-3D Generation via Coarse 3D Prior}
\author{Fangfu Liu$^{1}$, Diankun Wu$^{1}$, Yi Wei$^{1}$, Yongming Rao$^{2}$, Yueqi Duan$^{1\dagger}$\\
$^{1}$Tsinghua University, 
$^{2}$BAAI}

\begin{document}
\maketitle
\newcommand\blfootnote[1]{%
\begingroup 
\renewcommand\thefootnote{}\footnote{#1}%
\addtocounter{footnote}{-1}%
\endgroup 
}
\blfootnote{\textsuperscript{\dag}Corresponding author.}

\begin{strip}
    \centering
    \vspace{-6em}
    \centering
    \includegraphics[width=\textwidth]{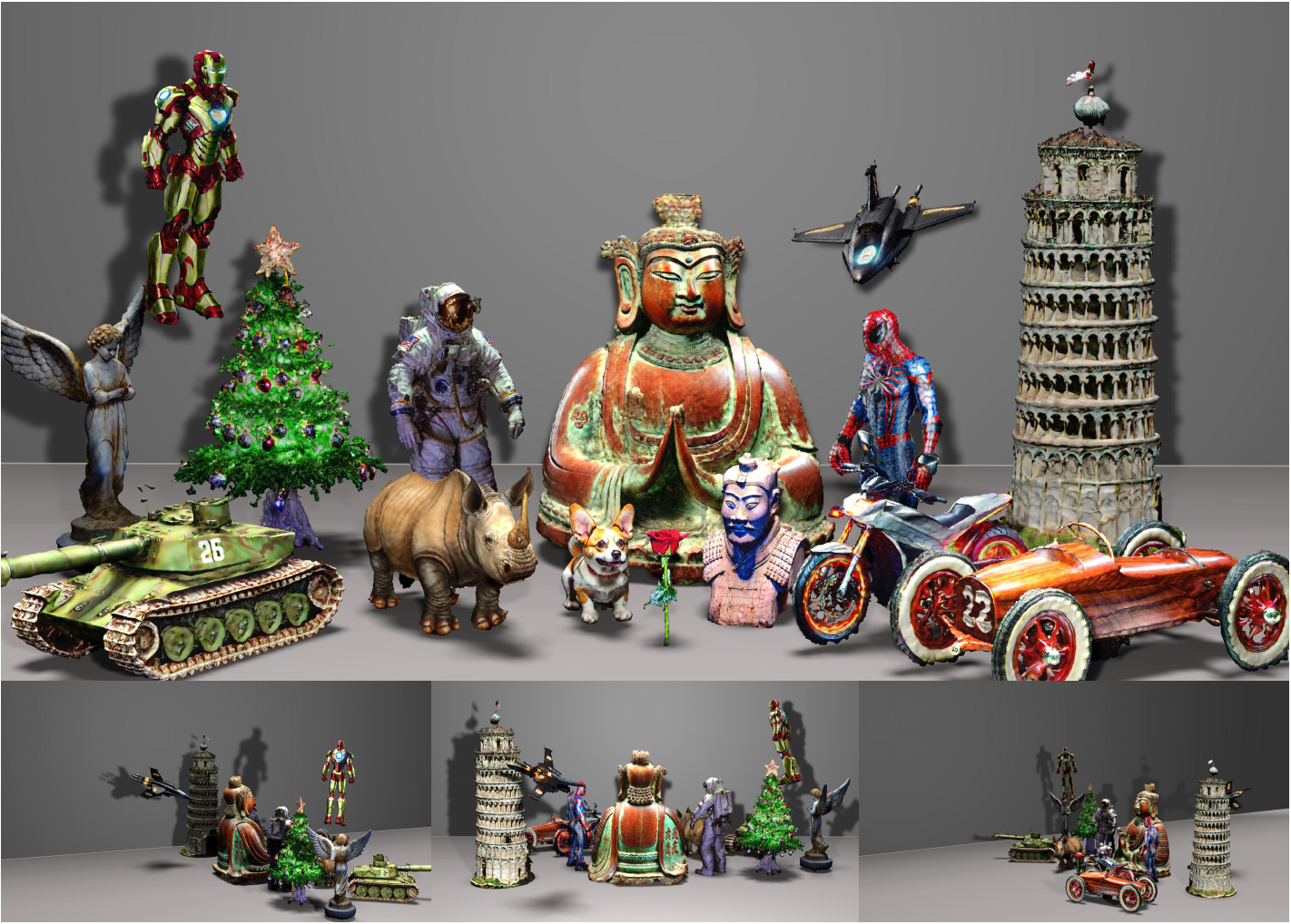}
    \vspace{-1em}
    \captionof{figure}{\textbf{Gallery of Sherpa3D:} Blender rendering for various textured meshes from \textit{Sherpa3D}, which is able to generate high-fidelity, diverse, and multi-view consistent 3D contents with input text prompts. Our method is also compatible with popular graphics engines.}
    \label{fig:teaser}
\end{strip}

\begin{abstract}
Recently, 3D content creation from text prompts has demonstrated remarkable progress by utilizing 2D and 3D diffusion models. 
While 3D diffusion models ensure great multi-view consistency, their ability to generate high-quality and diverse 3D assets is hindered by the limited 3D data. In contrast, 2D diffusion models find a distillation approach that achieves excellent generalization and rich details without any 3D data. 
However, 2D lifting methods suffer from inherent view-agnostic ambiguity thereby leading to serious multi-face Janus issues, where text prompts fail to provide sufficient guidance to learn coherent 3D results.
Instead of retraining a costly viewpoint-aware model, we study how to fully exploit easily accessible coarse 3D knowledge to enhance the prompts and guide 2D lifting optimization for refinement. In this paper, we propose \textbf{Sherpa3D}, a new text-to-3D framework that achieves high-fidelity, generalizability, and geometric consistency simultaneously. Specifically, we design a pair of guiding strategies derived from the coarse 3D prior generated by the 3D diffusion model: a structural guidance for geometric fidelity and a semantic guidance for 3D coherence. 
Employing the two types of guidance, the 2D diffusion model enriches the 3D content with diversified and high-quality results. Extensive experiments show the superiority of our Sherpa3D over the state-of-the-art text-to-3D methods in terms of quality and 3D consistency. Project page: \url{https://liuff19.github.io/Sherpa3D/}.
\end{abstract}

\section{Introduction}
3D content generation~\cite{li2023generative3d-aigc-survey,dreamfusion,wu2016learning,magic3d} finds a broad range of applications, including games, movies, virtual/augmented reality and robots. However, the conventional process of creating premium 3D assets is still expensive and challenging as it requires multiple labor-intensive and time-consuming stages~\cite{labschutz2011content}. Fortunately, this challenge has prompted the development of recent text-to-3D methods~\cite{dreamfusion,magic3d,jain2022zero-dreamfields,mohammad2022clip-mesh,chen2023fantasia3d, wang2023prolificdreamer, li2023sweetdreamer,hollein2023text2room, metzer2023latent-nerf}. Only using text prompts to automate 3D generation, these techniques pave a promising way towards streamlining 3D creation.

Powered by the great breakthroughs in diffusion models~\cite{stable-diffusion, imagen, nichol2022pointe, control-net}, two research lines of rationalization have recently emerged in text-to-3D: inference-only 3D diffusion methods and optimization-based 2D lifting methods. Specifically, the inference-only methods~\cite{jun2023shapE, nichol2022pointe, gupta20233dgen} seek to directly generate 3D-consistent assets by extensively training a new diffusion model on 3D data. However, 
due to the scarcity of 3D datasets compared to accessible 2D images or text data, these 3D diffusion models suffer from low quality and limited generalizability. Without requiring any 3D data for training, 2D lifting methods~\cite{dreamfusion, magic3d, metzer2023latent-nerf, chen2023fantasia3d, armandpour2023perp-neg, wang2023prolificdreamer, wang2022score-SJC} can produce high-quality and diversified 3D results by distilling 3D knowledge from pre-trained 2D diffusion models~\cite{stable-diffusion, imagen, control-net}, also known as Score Distillation Sampling (SDS). Yet lifting 2D observations into 3D is inherently ambiguous without sufficient 3D guidance from text prompts, leading to notorious multi-view inconsistency (\eg, Janus problems) in 2D lifting methods.

These findings motivate us to think: \textit{is it possible to bridge the two aforementioned streams to achieve generalizability, high-fidelity, and geometric consistency simultaneously?} An intuitive idea is to leverage more 3D data~\cite{deitke2023objaverse, deitke2023objaverse-xl} to fine-tune a view-point aware diffusion model, but it requires substantial computational resources and is prone to overfitting due to data bias~\cite{shi2023mvdream, liu2023zero123}. In contrast, our key insight is to utilize the easily accessible 3D diffusion model as guidance and study how to fully exploit coarse 3D knowledge to guide 2D lifting optimization for refinement.
In particular, when maintaining the quality and generalizability of the original 2D diffusion model, we hope the 2D lifting awareness can be guided by the strong 3D geometric information from the 3D diffusion model. However, it is non-trivial in pursuit of this balance. Relying too heavily on the coarse 3D priors from the 3D diffusion model may degrade the generation quality, whereas little 3D guidance could result in a lack of geometric awareness, leading to multi-view inconsistency. 

Towards this end, we propose \textbf{Sherpa3D} in this paper, which greatly boosts high-fidelity and highly diversified text-to-3D generation with geometric consistency. Our method begins by employing a 3D diffusion model to craft a basic 3D guide with limited details. Building upon the coarse 3D prior, we introduce two guiding strategies to inform 2D diffusion model throughout lifting optimization: a structural guide for geometric fidelity and a semantic guide for 3D coherence. Specifically, the structural guide leverages the first-order gradient information of the normals from the 3D prior to supervise the optimization of the structure. These normals are then integrated into the input of a pre-trained 2D diffusion model, refining the geometric details. Concurrently, our semantic guide extracts high-level features from multi-views of the 3D prior. These features guide the 2D lifting optimization to perceive the geometric consistency under the preservation of original generalizability and quality.
Furthermore, we design an annealing function, which modulates the influence of the 3D guidance to better preserve the capabilities of 2D and 3D diffusion models. As a result, our Sherpa3D  is aware of the geometric consistency 
with rich details and generalizes well across diverse text prompts. Extensive experiments verify the efficacy of our framework and show that our Sherpa3D outperforms existing methods for high-fidelity and geometric consistency (see qualitative results gallery in Figure~\ref{fig:teaser} and quantitative results in Table~\ref{tab:userstudy}).


\section{Related Work}
\subsection{Text-to-image Generation}
Recently, text-to-image models such as unCLIP~\cite{ramesh2022hierarchical-unclip}, Imagen~\cite{imagen}, and Stable Diffusion~\cite{stable-diffusion} have shown remarkable capability of generating high-quality and creative images given text prompts. Such significant progress is powered by advances in diffusion models~\cite{nichol2021improved-diffusion, song2020ddim, ho2020ddpm, dhariwal2021diffusion-beatgan}, which can be pre-trained on billions of image-text pairs~\cite{sharma2018conceptual, schuhmann2022laion-5B} and understands general objects with complex semantic concepts (nouns, artistic styles, \textit{etc}.)~\cite{stable-diffusion}. Despite the great success of photorealistic and diversified image generation, using language to generate different viewpoints of the same object with 3D coherence remains a challenging problem~\cite{watson2022novel-view}.

\subsection{Text-to-3D Generation}
Building on promising text-to-image diffusion models, there has been a surge of studies in text-to-3D generation. However, it is non-trivial due to the scarcity of diverse 3D data~\cite{chang2015shapenet, deitke2023objaverse, wu2023omniobject3d} compared to 2D. Existing 3D native diffusion models~\cite{jun2023shapE, nichol2022pointe, gupta20233dgen, lorraine2023att3d, zhang20233dshape2vecset, zheng2023locally} usually work on a limited object category and struggle with generating in-the-wild 3D assets. To achieve generalizable 3D generation, pioneering works DreamFusion~\cite{dreamfusion} and SJC~\cite{wang2022score-SJC} propose to distill the score of image distribution from pre-trained 2D diffusion models~\cite{stable-diffusion, imagen} and show impressive results. Following works~\cite{magic3d, tsalicoglou2023textmesh, zhu2023hifa, chen2023fantasia3d, yu2023points-to-3d, li2023focaldreamer, huang2023dreamtime, metzer2023latent-nerf, tang2023dreamgaussian, wang2023prolificdreamer} continue to enhance various aspects such as generation fidelity and optimization stability or explore more application scenarios~\cite{zhuang2023dreameditor, singer2023text-to-4d, raj2023dreambooth3d}. As it is inherently ambiguous to lift 2D observations into 3D, they may suffer from multi-face issues. Although some methods use prompt engineering~\cite{armandpour2023perp-neg} or train a costly viewpoint-aware model~\cite{liu2023zero123, shi2023mvdream} to alleviate such problems, they fail to generate high-quality results~\cite{chen2023fantasia3d} or easily overfit to domain-specific data~\cite{deitke2023objaverse, shi2023mvdream}.
In this work, we bridge the gap between 3D and 2D diffusion models through meticulously designed 3D guidance, which leads the 2D lifting process to achieve high-fidelity, diversified, and coherent 3D generation.

\subsection{3D Generative Models} 

Extensive research has been conducted in the field of 3D generative modeling, exploring diverse 3D representations like 3D voxel grids~\cite{gadelha20173d,henzler2019escaping,lunz2020inverse}, point clouds~\cite{achlioptas2018learning,luo2021diffusion,mo2019structurenet}, and meshes~\cite{gao2022get3d,zhang2020image}. The majority of these approaches rely on training data presented in the form of 3D assets, which proves challenging to obtain at a large scale. Drawing inspiration from the success of neural volume rendering, recent studies have shifted towards investing in 3D-aware image synthesis ~\cite{chan2022efficient,chan2021pi,gu2021stylenerf,hao2021gancraft,or2022stylesdf,schwarz2022voxgraf}. This approach offers the advantage of directly learning 3D generative models from images. However, volume rendering networks typically exhibit slow querying speeds, resulting in a trade-off between extended training times and a lack of multi-view consistency. Recently, benefitted from 2D diffusion models, some works generate multi-view images with single-view input~\cite{liu2023zero123,liu2023one,liu2023syncdreamer,shi2023zero123++,long2023wonder3d,yang2023consistnet}. As one of the pioneering works, Zero-1-to-3~\cite{liu2023zero123} uses a synthetic dataset to finetune the pretrained diffusion models, aiming to learn controls of the relative camera viewpoint. Beyond Zero-1-to-3, SyncDreamer~\cite{liu2023syncdreamer} employs a synchronized multiview diffusion model to capture the joint probability distribution of multiview images. This model facilitates the generation of multiview-consistent images through a unified reverse process. Different from these methods, we focus on text-to-3D synthesis, with the goal of generating multi-view consistent 3D contents with text prompts.

\section{Method}
\begin{figure*}
    \centering
    \includegraphics[width=1\textwidth]{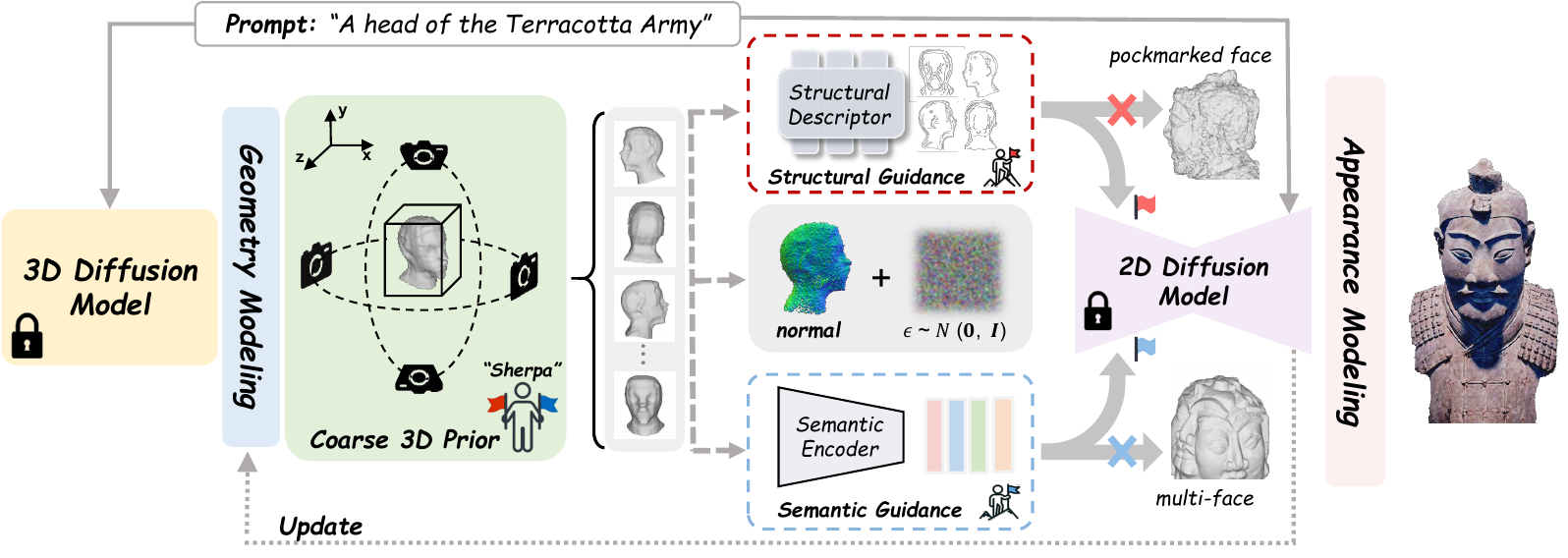}
    \caption{\textbf{Pipeline of our Sherpa3D.} Given a text as input, we first prompt 3D diffusion to build a coarse 3D prior $M$ encoded in the geometry model (\eg, DMTet). Next, we render the normal map of the extracted mesh in DMTet and derive two guiding strategies from $M$. (a) Structural Guidance: we utilize the structural descriptor to compute salient geometric features for preserving geometry fidelity (\eg, without a pockmarked face problem). (b) Semantic Guidance: we leverage a semantic encoder (\eg, CLIP) to extract high-level information for keeping 3D consistency (\eg, without multi-face issues). Employing the two guidance in 2D lifting process, we use the normal map as shape encoding of the 2D diffusion model and unleash its power to generate high-quality and diversified results with 3D coherence. Then we achieve the final 3D results via photorealistic rendering through appearance modeling. (\textit{``Everest's summit eludes many without \textbf{Sherpa}."})}
    \label{fig:pipeline}
    \vspace{-10pt}
\end{figure*}

Given a text prompt, our goal is to generate 3D assets with high quality, generalizability, and multi-view consistency. Our framework can be divided into three stages: (1) build coarse 3D prior from the 3D diffusion model (Sec.~\ref{sec: sculpt a 3D prior}); (2) formulate two guiding strategies (\eg, structural and semantic guidance) for 2D lifting process (Sec.~\ref{sec: 3D guidance}); (3) incorporate both 3D guidance and SDS loss with an annealing technique in optimization and generate the final 3D object (Sec.~\ref{sec: optimization}). In this way, we can leverage the full power of state-of-the-art 3D and 2D diffusion models to obtain 3D coherence as 3D models, retaining intricate details and creative freedom as 2D models. Our pipeline is depicted in Figure~\ref{fig:pipeline}. Before introducing our Sherpa3D in detail, we first review the theory of Score Distillation Sampling (SDS).

\subsection{Preliminaries}
\textbf{Score Distillation Sampling (SDS).} As one of the most representative 2D lifting methods, Dreamfusion~\cite{dreamfusion} first presents the concept of Score Distillation Sampling (SDS), which is an algorithm to optimize a 3D representation such that the image rendered from any view maintains a high likelihood as evaluated by the 2D diffusion model given text prompts. SDS consists of two key components: (1) a 3D representation with parameter $\theta$, which can produce an image $x$ at desired camera $\mathbf{c}$ through a parametric function $\mathbf{x}=g(\theta; \mathbf{c})$; (2) a pre-trained text-to-image 2D diffusion model $\phi$ with a score function $\epsilon_{\phi}(\mathbf{x}_t;y,t)$ that predicts the sample noise $\epsilon$ given noisy image $\mathbf{x}_t$, noise level $t$ and text embedding $y$. The score function guides the direction of the gradient for updating $\theta$ to reside rendered images in high-density areas conditioned on text $y$. The gradient is calculated by SDS as:
\begin{equation}
    \nabla_{\theta}\mathcal{L}_{\text{SDS}}(\phi, \mathbf{x}) = \mathbb{E}_{t, \epsilon}\left[w(t)\left(\epsilon_\phi\left(\mathbf{x}_t ; y, t\right)-\epsilon\right) \frac{\partial \mathbf{x}}{\partial \theta}\right],
    \label{eq: sds-loss}
\end{equation}
where $w(t)$ is a weighting function. In practice, the denoising score function $\epsilon_{\phi}$ is often replaced with another function $\tilde{\epsilon}_\phi$ that uses classifier-free guidance~\cite{ho2022classifier-guidance} that controls the strength of the text condition (see Supplementary).

\subsection{Sculpting a Coarse 3D Prior}
\label{sec: sculpt a 3D prior}
To facilitate text-to-3D generation, most existing methods~\cite{dreamfusion,metzer2023latent-nerf, huang2023dreamtime} rely on implicit 3D representations such as Neural Radiance Fields (NeRF)~\cite{mildenhall2020nerf} and its variants~\cite{barron2021mip-nerf, muller2022instant-ngp}. However, it is difficult for NeRF-based modeling to extract the high-quality surface with material and texture~\cite{wang2021neus}. 
To address this, we adopt the hybrid scene representation of DMTet~\cite{shen2021dmtet}, including a deformable tetrahedral grid that encodes a signed distance function (SDF) and a differentiable marching tetrahedra (MT) layer that extracts explicit surface mesh. Equipped with the hybrid representation, we sculpt a coarse 3D prior from 3D diffusion model $G_{3D}$ (\eg, Shap-E~\cite{jun2023shapE}) by the following procedure. Given text prompts $y$, we first use the 3D diffusion model $G_{3D}$ to generate 3D results $M_{0}$ and employ multi-layer perceptions (MLPs) to query SDF values for each vertex along a regular grid. Next we sample a point set $\mathcal{P}=\{ {\boldsymbol{p}_i \in \mathbb{R}^3} \}$ from $M_0$ with their SDF values $\{ SDF(\boldsymbol{p}_i) \}$. For each $\boldsymbol{p}_i$, the DMTet network $\mathcal{F}$ can predict SDF value $s(\boldsymbol{p}_i), $ and a position offset $\Delta \boldsymbol{p}_i$ by:
\begin{equation}
    (s(\boldsymbol{p}_i), \Delta \boldsymbol{p}_i) = \mathcal{F}(\boldsymbol{p}_i; \theta),
\end{equation}
where $\theta$ is the parameters of network $\mathcal{F}$. Then, we incorporate 3D priors into the DMTet network $\mathcal{F}_{\theta}$ with the point set derived from 3D diffusion model by minimizing:
\begin{equation}
\small
    \mathcal{L}_{SDF} = \sum_{\boldsymbol{p}_i \in \mathcal{P}} |s(\boldsymbol{p}_i) - SDF(\boldsymbol{p}_i)|^2 + \lambda_{def} \sum_{\boldsymbol{p}_i \in \mathcal{P}} ||\Delta  \boldsymbol{p}_i||_2,
\end{equation}
\normalsize 
where $\lambda_{def}$ is the hyperparameter controlling $L_2$ regularization strengths on offsets to avoid artifacts.
Finally, we apply the MT layer to extract mesh representation $M$. Now, we have leveraged the knowledge from the 3D diffusion model to construct a coarse 3D prior, which is encoded implicitly in DMTet $\mathcal{F}_\theta$ and represented explicitly by mesh $M$. Next, we will discuss how to utilize the coarse 3D prior $M$ as a guide during the subsequent 2D diffusion lifting optimization to refine a high-quality result with 3D coherence.

\subsection{3D Guidance for 2D Lifting Optimization}
\label{sec: 3D guidance}
\noindent \textbf{What knowledge can serve as guidance?} 
The purpose of introducing a 3D prior as guidance is to address the prevalent issue of viewpoint inconsistency both in geometry and appearance. Through empirical studies, we have identified geometric inconsistency as the main cause of 3D incoherence, leading to multi-face Janus problem~\cite{shi2023mvdream,li2023sweetdreamer}. In contrast, appearance inconsistency emerges in much more extreme scenarios with lesser significance.
Therefore, we disentangle the geometry from the 3D model and fully leverage coarse prior $M$ to guide 2D lifting geometry optimization with view-point awareness. Our analysis of the coarse 3D prior indicates that it contains the essential geometric structures and captures the basic categorical attributes, keeping semantic rationality across different views. Building upon these observations, a natural insight is to preserve such inherent 3D knowledge as guidance and continuously benefit the 2D lifting process.
For example, given text prompts ``a head of the Terracotta Army," we hope the knowledge in the guidance can prevent issues such as a pockmarked face or the unrealistic scenario of having a face on the back (\eg, Janus problem). 
To this end, we have designed two guiding strategies derived from $M$: structural guidance for geometric fidelity and semantic guidance for 3D coherence. 

\noindent \textbf{Structural guidance.}
Given the current DMTet net $\mathcal{F}$ with parameters $\theta$ that encodes the coarse 3D prior $M$, we apply a differentiable render $f_n$ (\eg, nvidiffrast~\cite{laine2020nviddiffrast}) to generate a set of normal maps $\mathcal{N} =\{{\boldsymbol{n}_i | \boldsymbol{n}_i = f_n(\mathcal{F}_{\theta}, \mathbf{c}_i)}, i=1,...n\}$, where $\mathbf{c}_i$ is the camera position randomly sampled in spherical coordinates. To extract the salient geometric structure features, we first use a Gaussian filter with a kernel standard deviation $\sigma$
\begin{equation}
    G(x, y)=\frac{1}{2 \pi \sigma^2} e^{-\frac{x^2+y^2}{2 \sigma^2}}
\end{equation}
to reduce the noise impact and obtain $\{\sigma(\boldsymbol{n}_i) \}$. As gradients are simple but effective tools for revealing the geometric contours and salient structures~\cite{ding2001canny, kanopoulos1988design-sobel}, we then compute the structural descriptor sets $\{G_{\sigma}(\boldsymbol{n}_i)\}$ by
\begin{equation}
    G_{\sigma}(\boldsymbol{n}_i) = \sqrt{(\frac{\partial \sigma(\boldsymbol{n}_i)}{\partial x})^2 + (\frac{\partial \sigma(\boldsymbol{n}_i)}{\partial y})^2},
\end{equation}
where $x$ and $y$ are the coordinate directions of the normal map $\boldsymbol{n}_i$. 
Throughout the 2D lifting process of updating $\mathcal{F}_{\theta}$ with newly rendered normal maps $ \tilde{\mathcal{N}}=\{ \tilde{\boldsymbol{n}}_i\}$, it should follow the structural guidance as:
\begin{equation}
    \min_{\theta} \mathcal{L}_{\text{struc}} := \sum_{i=1}^n ||G_{\sigma}(\boldsymbol{n}_i) - G_{\sigma}(\boldsymbol{\tilde{n}}_i)||^2_2,
\end{equation}
which enables the 2D lifting process to preserve geometric fidelity and a well-aligned structure with the coarse 3D prior when generating rich details.

\noindent \textbf{Semantic guidance.} 
 While structural guidance maintains low-level geometric perception from coarse 3D prior, semantic guidance extracts high-level features for 3D coherence. We first apply the pre-trained CLIP~\cite{radford2021-CLIP} model as semantic encoder $\psi$ to the normal set $\mathcal{N}$ and obtain semantic feature maps $\mathcal{N}_c = {\{\psi(\boldsymbol{n}_i)\}}$, proven to effectively capture semantic attributes like facial expressions or view categories~\cite{goh2021multimodal}. Following the notation as above, we then define the semantic guidance with cosine similarity:
 \begin{equation}
     \min_{\theta} \mathcal{L}_{\text{sem}} := \sum_{i=1}^n \frac{\psi(\boldsymbol{n}_i)\cdot \psi(\boldsymbol{\tilde{n}}_i)}{\| \psi(\boldsymbol{n}_i)\|\|\psi(\boldsymbol{\tilde{n}}_i)\|}.
 \end{equation}
Employing this guidance, we ensure that different views retain inherent high-level information throughout the 2D lifting optimization process. Experiments show that it can effectively mitigate multi-face problems, keeping 3D content semantically plausible from all viewing angles.

\subsection{Optimization}
\label{sec: optimization}
In this subsection, we incorporate both structural and semantic guidance derived from coarse 3D prior to 2D lifting optimization so that it can produce vivid and diversified objects with multi-view consistency. For the disentangled geometry modeling, we use the randomly sampled normal map $\boldsymbol{n}$ as the input, bridging the gap between 3D and 2D diffusion. To update the geometry model DMTet network $\mathcal{F}_\theta$, we choose to use the publicly available \textit{Stable Diffusion}~\cite{stable-diffusion} as pre-trained 2D diffusion model $\phi$ and compute the gradient of the SDS loss similar in Eq.~\ref{eq: sds-loss}:
\begin{equation}
\small
    \nabla_{\theta}\mathcal{L}_{\text{SDS}}(\theta, \boldsymbol{n}) =\mathbb{E}_{t, \epsilon}\left[w(t)\left(\epsilon_\phi\left(\boldsymbol{z}_t^{\boldsymbol{n}} ; y, t\right)-\epsilon\right) \frac{\partial \boldsymbol{z}_t^{\boldsymbol{n}}}{\partial \boldsymbol{n}} \frac{\partial \boldsymbol{n}}{\partial \theta}\right],
    \label{eq: geo-sds}
\end{equation}
\normalsize
where ${\partial \boldsymbol{z}_t^{\boldsymbol{n}}}/{\partial \boldsymbol{n}}$ calculates the gradient of the encoder in the latent diffusion model (LDM)~\cite{stable-diffusion}. Additionally, we introduce a step annealing technique to balance the influence of the 3D guidance during 2D lifting optimization:
\begin{equation}
    \gamma(\lambda) = \lambda  e^{-\beta \max (0, n_{\text{cur}} - m)},
\end{equation}
where $n_{\text{cur}}$ is the current epoch and $\{\beta, m ,\lambda \}$ are the hyperparameters that control how $\gamma$ decreased. Therefore, the total loss $\mathcal{L}_{\text{geo}}$ to lift 2D geometry optimization with 3D guidance is a weighted sum of three loss terms:
\begin{equation}
    \mathcal{L}_{\text{geo}}(\theta,\boldsymbol{n}) = \mathcal{L}_{\text{SDS}} + \gamma(\lambda_{\text{struc}})\mathcal{L}_{\text{struc}} + \gamma(\lambda_{\text{sem}})\mathcal{L}_{\text{sem}}, 
\end{equation}
which not only enables the 3D content generation without multi-view inconsistency issues but also preserves the generalization and quality in 2D diffusion model $\phi$. As our pipeline can be integrated into any appearance model~\cite{chen2023text2tex, lei2022tango, chen2023fantasia3d}, we adopt a similar approach as Fantasia3D~\cite{chen2023fantasia3d} to better align our text and 3D object. Denote $\mathcal{T}$ with parameters $\eta$ as our appearance model, we have the rendered image $\mathbf{x} = \mathcal{T}_{\eta}(\mathcal{F}_\theta, \boldsymbol{c}_i)$. To update $\eta$, we again apply the SDS loss for the final complete generated 3D object with detailed texture and coherent geometry:
\begin{equation}
    \small
    \nabla_{\eta}\mathcal{L}_{\text{app}}(\eta, \mathbf{x}) =\mathbb{E}_{t, \epsilon}\left[w(t)\left(\epsilon_\phi\left(\boldsymbol{z}_t^{\mathbf{x}} ; y, t\right)-\epsilon\right) \frac{\partial \boldsymbol{z}_t^{\mathbf{x}}}{\partial \mathbf{x}} \frac{\partial \mathbf{x}}{\partial \eta}\right],
\end{equation}
which shares similar notations defined in Eq.~\ref{eq: geo-sds}. Finally, through the tailored 3D structural and semantic guidance that bridges the 2D and 3D diffusion models, our Sherpa3D can mitigate the multi-face problem and achieve high-fidelity and diversified results.

\subsection{Implementation Details}
We apply the multilayer perceptron (MLP) comprising of three hidden layers to approximate $\mathcal{F}_{\theta}$ and $\mathcal{T}_{\eta}$. Adam optimizer~\cite{kingma2014adam} is used to update $\mathcal{F}_\theta$ and $\mathcal{T}_{\eta}$ with an initial learning rates of $1e-3$ decaying into $5e-4$.
For 3D representations, we use textured mesh with a DMTet resolution of 128 to achieve a balance between quality and generation speed. 
We sample random camera poses at a fixed radius of $2.5$, y-axis FOV of $45^{\circ}$, with the azimuth in $[-180^{\circ}, 180^{\circ}]$ and elevation in $[-30^{\circ}, 30^{\circ}]$. We load Shap-E from~\cite{luo2023scalable-shape} for 3D diffusion model and choose \textit{stabilityai/stable-diffsuion-2-1-base}~\cite{stable-diffusion} for 2D diffusion model. For weighting factors, we follow the same strategy as~\cite{huang2023dreamtime} to tune $w(t)$. 
$\lambda_{\text{struc}}$ is set to $10$ and $\lambda_{\text{sem}}$ is $30$ to balance the magnitude of SDS loss.
Notably, our method only needs a single NVIDIA RTX3090 (24GB) GPU within 25 minutes.
More details of optimization, architecture design, and hyperparameter settings can be found in the supplementary.

\begin{figure*}[!t]
    \centering
    \includegraphics[width=0.90\linewidth]{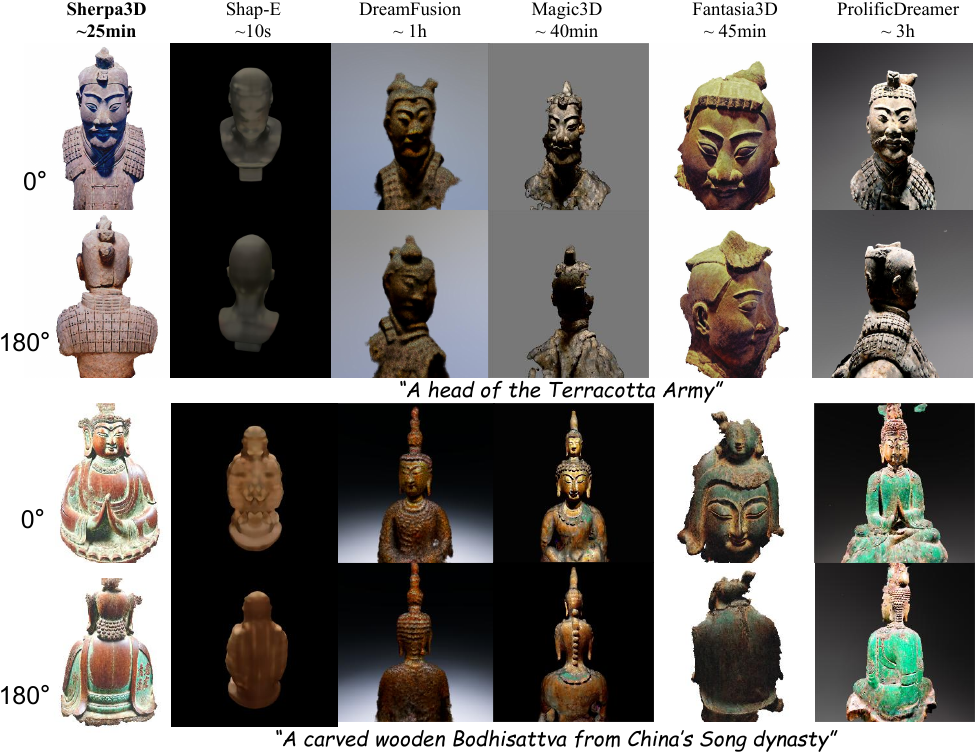}
    \caption{\textbf{Qualitative comparisons} with baseline methods across different views ($0^{\circ}$ and $180^\circ$). We can observe that baseline methods suffer from severe multi-face issues while our Sherpa3D can achieve better quality and 3D coherence. }
    \label{fig:visualation_up}
\end{figure*}

\section{Experiments}
In this section, we conduct comprehensive experiments to evaluate our text-to-3D framework Sherpa3D and show comparison results against other text-to-3D baseline methods. We first present qualitative results compared with five SOTA baselines from different viewpoints. Then we report the quantitative results with a user study. Finally, we carry out ablation studies to further verify the efficacy of our framework design. Please refer to the supplementary for more comparisons, visualizations, and detailed analysis.

\subsection{Experiment Setup}
\noindent \textbf{Baselines.} We extensively compare our method Sherpa3D against five baselines: Shap-E~\cite{jun2023shapE}, DreamFusion~\cite{dreamfusion}, Magic3D~\cite{magic3d}, ProlificDreamer~\cite{wang2023prolificdreamer}, and Fantasis3D~\cite{chen2023fantasia3d}. Due to various reasons, we can't obtain the original implementation of some baselines. For DreamFusion, Magic3D, and ProlificDreamer, we utilize their implementations in the Threestudio library~\cite{threestudio2023} for comparison. For Shap-E and Fantasia3D, we follow their official implementation. We consider these implementations to be the most reliable and comprehensive open-source option available in the field. To ensure a fair comparison, we use the Stable Diffusion~\cite{stable-diffusion} model as 2D diffusion prior by default.

\noindent
\textbf{Metrics.} We will show our results with notable comparisons to other baselines through visualization. As there is no Ground-Truth 3D content corresponding to the text prompt, reference-based metrics like Chamfer Distance are difficult to apply to zero-shot text-to-3D generation. Following~\cite{dreamfusion, jain2022zero-dreamfields}, we evaluate the CLIP R-Precision~\cite{park2021benchmark}, which can measure how well the rendered images of generated 3D content align with the input text. We use 100 prompts from the Common Objects in Context (COCO) dataset~\cite{lin2014microsoft-coco} as DreamFusion~\cite{dreamfusion}. we also conduct a user study to further demonstrate the multi-view consistency and overall generation quality of our method,

\begin{figure*}[!t]
    \centering
    \includegraphics[width=0.90\linewidth]{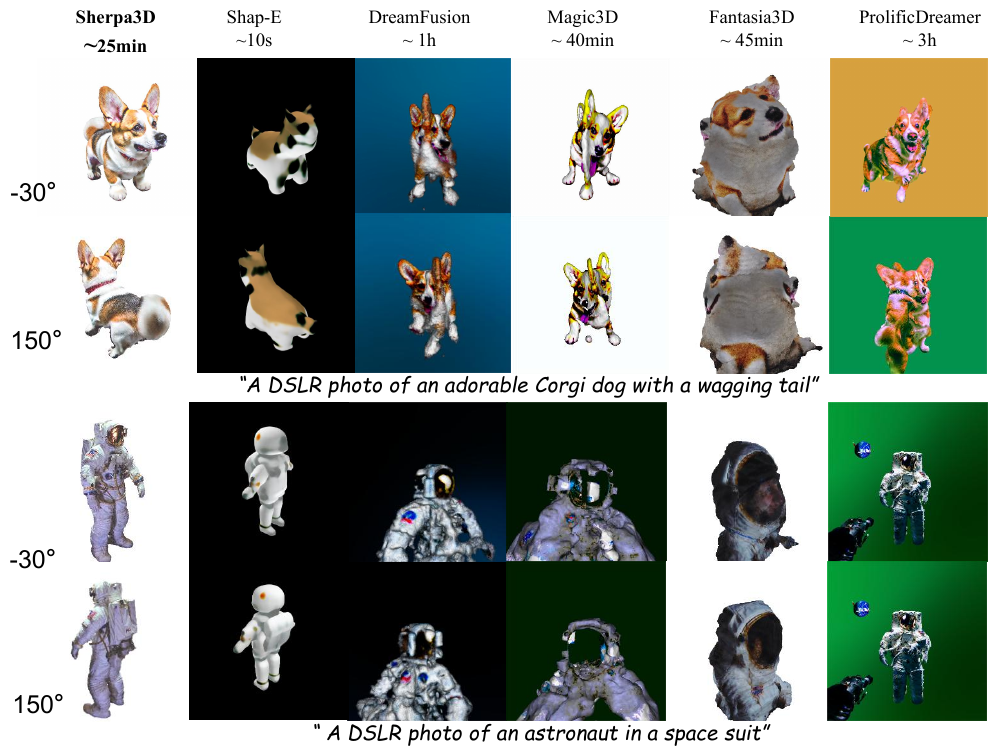}
    \caption{\textbf{Qualitative comparisons} with baseline methods across different views ($-30^{\circ}$ and $150^\circ$).}
    \label{fig:visualation_down}
\end{figure*}

\subsection{Qualitative Comparisons}
We first demonstrate vivid and diversified text-to-3D results generated from our Sherpa3D in the gallery as shown in Figure~\ref{fig:teaser}. Then we compare our method with five baseline method: Shap-E~\cite{jun2023shapE}, DreamFusion~\cite{dreamfusion}, Magic3D~\cite{magic3d}, Fantasia3D~\cite{chen2023fantasia3d} and ProlificDreamer~\cite{wang2023prolificdreamer}. Figure~\ref{fig:visualation_up} and~\ref{fig:visualation_down} give the comparative results with the same text prompt for each object generation. We observe that the Shap-E~\cite{jun2023shapE} only generates coarse shapes while other 2D lifting methods suffer from multi-face problems. In contrast, our Sherpa3D produces high-fidelity 3D assets with compelling texture quality and multi-view consistency. Notably, our framework is more efficient than other baselines with less time to optimize. Specifically, it only takes within 25 minutes from a text prompt to a high-quality 3D model ready to be used in graphic engines.


\subsection{Quantitative Comparisons}
In Table~\ref{tab:rp}, we report the CLIP R-Precision for Sherpa3D and several baselines. It shows that our method outperforms other baselines consistently across different CLIP models, and approaches the performance of ground truth (GT) images. For the user study, we render 360-degree rotating videos of 3D models generated from a collection of 120 images. Each volunteer is shown 10 samples of rendered video from a random method and rates in two aspects: multi-view consistency and overall generation quality. We collect results from 50 volunteers shown in Table~\ref{tab:userstudy}. We observe that most users consider our results with much higher viewpoints consistency and overall generation fidelity. 

\begin{table}[t]
\caption{\textbf{Quantitative comparisons} on generation renderings with text prompts using different CLIP retrieval models. We compared to ground-truth images, Shap-E~\cite{jun2023shapE}, Dreamfusion~\cite{dreamfusion}, Magic3D~\cite{magic3d}, evaluated on object-centric COCO as in~\cite{dreamfusion}.}

\resizebox{\linewidth}{!}{
\begin{tabular}{cccc}
\hline \multirow{2}{*}{ Method } & \multicolumn{3}{c}{ R-Precision (\%) $\uparrow$} \\
& CLIP B/32 & CLIP B/16 & CLIP L/14 \\
\hline GT Images & 77.3 & 79.2 & - \\
\hline Shape-E~\cite{jun2023shapE} & 41.1 & 42.5 & 46.4 \\
DreamFusion~\cite{dreamfusion} & 70.3 & 73.2 & 75.0 \\
Magic3D~\cite{magic3d} & 71.5 & 73.8 & 76.1 \\
Sherpa3D (Ours) & \textbf{72.3} & \textbf{75.6} & \textbf{79.3} \\
\hline
\end{tabular}}
\label{tab:rp}
\end{table}

\begin{figure*}[!t]
    \centering
    \includegraphics[width=1\linewidth]{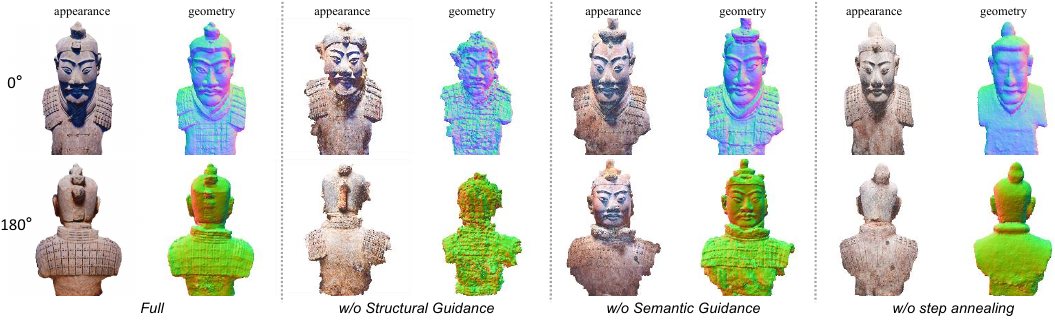}
    \caption{\textbf{Ablation study of our method.} The generation is based on the text prompt ``a head of the Terracotta Army". We ablate the design choices of structural guidance, semantic guidance (Sec.~\ref{sec: 3D guidance}), and the step annealing technique (Sec.~\ref{sec: optimization}).}
    \label{fig:ablation}
\end{figure*}

\begin{table}[!t]
\caption{\textbf{Quantitative comparisons} on the multi-view consistency and overall generation quality score in a user study, rated on a scale of 1-10, with higher scores indicating better performance.}
\resizebox{\linewidth}{!}{
\begin{tabular}{ccc}
\hline
Method                  & Multi-view Consistency $\uparrow$ & Overall Quality $\uparrow$ \\ \hline
Shap-E~\cite{jun2023shapE}                 & 6.09                   & 3.33             \\
DreamFusion~\cite{dreamfusion}             & 4.58                   & 6.05            \\
Magic3D~\cite{magic3d}                 & 5.25                   & 6.73             \\
Fantasia3D~\cite{chen2023fantasia3d}              & 3.83                   & 5.90             \\
ProlificDreamer~\cite{wang2023prolificdreamer}         & 5.78                   & 7.02             \\
\textbf{Sherpa3D(Ours)} & \textbf{8.95}          & \textbf{8.74}    \\ \hline
\end{tabular}}
\label{tab:userstudy}
\end{table}

\subsection{Ablation Study and Analysis}
We carry out ablation studies on the design of our Sherpa3D framework in Figure~\ref{fig:ablation} using an example text prompt ``a head of the Terracotta Army". Specifically, we perform ablation on three aspects of our method: structural guidance, semantic guidance, and the step annealing strategy. The results reveal that the omission of any of these elements leads to a degradation in terms of quality and consistency. Notably, the absence of structural guidance leads to a loss of geometric fidelity in the ``army'', leading to a pockmarked face; without semantic guidance, there's a loss of semantic rationality across different views, resulting in the multi-view Janus problem. The lack of a balanced step annealing results in an excessive influence of guidance with a rough final output. This illustrates the effectiveness of our overall framework (Figure~\ref{fig:pipeline}), which drives geometric fidelity, multi-view consistency, and optimization balance steered by the 3D guidance and annealing strategy.

\begin{figure}
    \centering
    \includegraphics[width=1\linewidth]{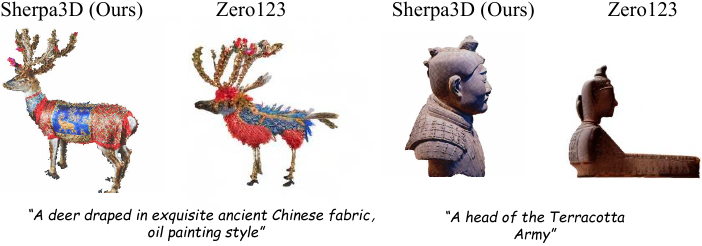}
    \caption{Comparison with Zero123~\cite{liu2023zero123}. We use the front view of our generated 3D model as the input of Zero123 with open-vocabulary text prompts.}
    \label{fig:zero123}
\end{figure}

To further demonstrate our generalizability, we compare our method in Figure~\ref{fig:zero123} with the Zero123~\cite{liu2023zero123} which uses more 3D data~\cite{liu2023zero123} to finetune a 2D diffusion model to be viewpoint-aware. However, such a finetuning-based method easily overfits to 3D training data and suffers from severe performance degradation with unseen input of the training set. In contrast, our method is more generalizable to open-vocabulary text prompts.

\section{Conclusion}

In this paper, we present Sherpa3D, a new framework that simultaneously achieves high-quality, diversified, and 3D consistent text-to-3D generation. By fully exploiting easily obtained coarse 3D knowledge from the 3D diffusion model, we derive structural guidance and semantic guidance to enhance the prompts and provide continuous guidance with geometric fidelity and 3D coherence throughout the 2D lifting optimization. To further improve the overall performance, we incorporate a step annealing strategy that modulates the impact of 3D guidance and 2D refinement. 
Therefore, our framework bridges the gap between 2D and 3D diffusion models, preserving multi-view coherent generation while maintaining the generalizability and fidelity of 2D models.
Extensive qualitative and quantitative experiments verify the remarkable improvement of our Sherpa3D on text-to-3D generation. 
\newline
\newline
\noindent \textbf{Limitations and future works.} Although our Sherpa3D achieves remarkable text-to-3D results, the quality still seems to be limited to the backbone itself as we choose Shap-E~\cite{jun2023shapE} and Stable Diffusion v2.1 base model in this work. We expect them to be solved with a larger diffusion model, such as SDXL~\cite{SDXL} and DeepFloyd~\cite{deepfloyd}. In future work, we are interested in extending our insight to more creative text-to-4D generation. We believe that Sherpa3D provides a promising research path for user-friendly and more accessible 3D content creation.

{
    \small
    \bibliographystyle{ieeenat_fullname}
    \bibliography{cvpr2024_conference}
}

\clearpage
\maketitlesupplementary

\section{More Discussion of Preliminaries}
In this section, we provide more preliminaries and details of our implementation for Score Distillation Sampling (SDS).

\subsection{Diffusion Models}
The diffusion model, which is a type of likelihood-based generative model used to learn data distributions, has been studied extensively in recent years~\cite{sohl2015deep-diffusion, song2020score-based-diffusion, song2019generative, ho2020ddpm, song2020ddim}. Given an underlying data distribution $q_0(\boldsymbol{x})$, a diffusion model composes two processes: (a) a forward process $\{q_t \}_{t \in [0,1]}$ to gradually add noise to the data point $\boldsymbol{x}_0 \sim q_0(\boldsymbol{x}_0)$; (b) a reverse process $\{p_t \}_{t\in [0,1]}$ to denoise data (\eg, generation). Specifically, the forward process is defined by $q_t\left(\boldsymbol{x}_t \mid \boldsymbol{x}_0 \right):=\mathcal{N}\left(\alpha_t \boldsymbol{x}_0 , \sigma_t^2 \boldsymbol{I}\right)$ and $q_t\left(\boldsymbol{x}_t\right):=\int q_t\left(\boldsymbol{x}_t \mid \boldsymbol{x}_0\right) q_0\left(\boldsymbol{x}_0\right) \mathrm{d} \boldsymbol{x}_0$, where $\alpha_t, \sigma_t > 0$ are hyperparameters. On the other hand, the reverse process is described with the transition kernel $p_t(\boldsymbol{x}_{t-1} \mid \boldsymbol{x}_t) := \mathcal{N} (\mu_{\phi} (\boldsymbol{x}_t, t), \sigma_t^2 \boldsymbol{I})$ from $p_1(\boldsymbol{x}_1) := \mathcal{N}(\boldsymbol{0}, \boldsymbol{I})$. The training objective is to optimize $\mu_{\phi}$ by maximizing a variational lower bound of a log-likelihood. In practice, $\mu_{\phi}$ is re-parameterized as a denoising network $\boldsymbol{\epsilon}_{\phi}(\boldsymbol{x}_t, t)$~\cite{ho2020ddpm} to predict the noise added to the clean data $\boldsymbol{x}_0$, which is trained by minimizing the MSE criterion~\cite{ho2020ddpm, kingma2021variational}:
\begin{equation}
\small
    \mathcal{L}_{\text {Diff }}(\phi):=\mathbb{E}_{\boldsymbol{x}_0, t, \boldsymbol{\epsilon}}\left[\omega(t)\left\|\boldsymbol{\epsilon}_\phi\left(\alpha_t \boldsymbol{x}_0+\sigma_t \boldsymbol{\epsilon}\right)-\boldsymbol{\epsilon}\right\|_2^2\right],
\end{equation}
\normalsize
where $\omega(t)$ is the time-dependent weights. Besides, the noise prediction network $\boldsymbol{\epsilon}_{\phi}$ can be applied for approximating the score function~\cite{song2019generative} of the perturbed data distribution $q(\boldsymbol{x}_t)$, which is defined as the gradient of the log-density:
\begin{equation}
    \nabla_{\boldsymbol{x}_t} \log q_t\left(\boldsymbol{x}_t\right)  \approx-\boldsymbol{\epsilon}_\phi\left(\boldsymbol{x}_t, t\right) / \sigma_t .
\end{equation}
This means that the diffusion model can estimate a direction that guides $\boldsymbol{x}_t$ towards a high-density region of $q(\boldsymbol{x}_t)$, which is the key idea Score Distillation Sampling (SDS)~\cite{wang2022score-SJC, dreamfusion} for optimizing the 3D scene via well 2D pre-trained models. 

\subsection{SDS with Classifier-Free Guidance}
As one of the most successful applications of diffusion models, text-to-image generation~\cite{stable-diffusion, imagen, ramesh2022hierarchical-unclip} generate samples $\boldsymbol{x}$ based on the text prompt $y$, which is also fed into the $\boldsymbol{\epsilon}_{\phi}$ as input, denoted as $\boldsymbol{\epsilon}_\phi\left(\boldsymbol{x}_t ; t, y\right)$. An important technique to improve the performance of these models is Classifier-Free Guidance (CFG)~\cite{ho2022-CFG}. CFG modifies the original model by adding a guidance term, \ie, $\hat{\boldsymbol{\epsilon}}_{\phi}(\boldsymbol{x}_t; y,t):= (1+s) \boldsymbol{\epsilon}_{\phi}(\boldsymbol{x}_t; y, t) - s \boldsymbol{\epsilon}_{\phi} (\boldsymbol{x}_t;t, \varnothing) $, where $s > 0$ is the guidance weight that controls the balance between fidelity and diversity, while $\varnothing$ denotes the ``empty'' text prompt for the unconditional case. Recall the SDS gradient form to update $\theta$: 
\begin{equation}
    \nabla_{\theta}\mathcal{L}_{\text{SDS}}(\phi, \boldsymbol{x}) = \mathbb{E}_{t, \epsilon}\left[\omega(t)\left(\boldsymbol{\epsilon}_\phi\left(\boldsymbol{x}_t ; y, t\right)-\epsilon\right) \frac{\partial \boldsymbol{x}}{\partial \theta}\right],
    \label{eq: sds-loss}
\end{equation}
and denote $\delta_{\boldsymbol{x}}(\boldsymbol{x}_t; y,t) := \epsilon_{\phi}(\boldsymbol{x}_t;y,t) -  \boldsymbol{\epsilon}$. In principle, $\boldsymbol{\epsilon}(\boldsymbol{x}_t; y, t)$ should represent the pure text-conditioned score function in Eq.~(\ref{eq: sds-loss}). But in practice, CFG is employed in it with a guidance weight $s$ to achieve high-quality results, where we rewrite
\begin{equation}
\small
    \delta_{\boldsymbol{x}}(\boldsymbol{x}_t; y, t) = [\boldsymbol{\epsilon}_{\phi}(\boldsymbol{x}_t; y, t) - \boldsymbol{\epsilon}] + s[\boldsymbol{\epsilon}_{\phi}(\boldsymbol{x}_t; y,t) - \boldsymbol{\epsilon}_{\phi}(\boldsymbol{x}_t; t, \varnothing)].
    \label{eq: sds-cfg}
\end{equation} 
As DreamFusion~\cite{dreamfusion} uses $s=100$ for high fidelity, our implementation adopts $s=50$ with the enhancement of structural and semantic guidance to preserve some diversity. The two types of guidance can also be seen as another form of prompt guidance that is more generalizable and robust.
Therefore, there is a gap between the original formulation in Eq.~(\ref{eq: sds-loss}) and the practical coding implementation in Eq.~(\ref{eq: sds-cfg}). 

\section{Additional Implementation Details}
\noindent \textbf{Training details.} Our geometry model $\mathcal{F}_{\theta}$ and appearance model $\mathcal{T}_{\eta}$ is approximated by three-layer MLPs and we apply adam~\cite{kingma2014adam} optimizer to update them with an initial learning rates of $1\times 10^{-3}$ to decaying to $5 \times 10^{-4}$. In particular, our method is optimized for 2500 iterations about 15 minutes to learn $\mathcal{F}_{\theta}$ and 2500 iterations about 10 minutes to learn $\mathcal{T}_{\eta}$. For geometry modeling, we utilize the Open3D library~\cite{zhou2018open3d} to calculate the signed distance function (SDF) value for each point in Equations 2 and 3 in the main paper. In our experiments, the DMTet-based coarse 3D prior building stage is critical as it not only provides coarse 3D knowledge with consistency but also boosts the speed of the convergence of generation. For appearance modeling, since our focus in this paper is to fully exploit easily obtained coarse 3D knowledge that serves as guidance for 2D lifting optimization (as discussed in Section 3.3 of our paper), we do not design a specific appearance model for our framework. Note that our geometry model is plug and play and we can leverage different models~\cite{chen2023text2tex, chen2023fantasia3d, lei2022tango}, we leverage the same PBR materials approach in Fantasia3D~\cite{chen2023fantasia3d} to achieve photorealistic surface renderings and better aligns with our geometry modeling.

\noindent \textbf{Hyperparameter settings.} We select the camera positions $(r, \kappa, \varphi)$ in the spherical coordinate system, where $r$ denote radius, $\kappa$ is the elevation and $\varphi$ is the azimuth angle respectively. Specifically, we sample random camera poses at a fixed $r=2.5$ with the $\kappa \in [-30^{\circ}, 30^{\circ}]$. In a batch of $b \times l$ images, we partition $\varphi$ into $l$ intervals in $[-180^{\circ}, 180^{\circ}]$ and uniformly sample $b$ azimuth angles in each interval. 
For structural guidance, we set $\sigma=1$ in Eq.~(4) in the main paper as the standard deviation of the Gaussian filter. We tune $\lambda_{\text{struc}}$ and $\lambda_{\text {sem}}$ in $\{0.01, 0.1, 1, 5, 10, 20, 30, 100\}$. We find that often $\lambda_{\text{struc}} = 10$ and $\lambda_{\text {sem}}=30$ works well with $\beta =0.5$ in the step annealing technique, which may balance the magnitude of SDS losses and better guide the 2D lifting to refine the 3D contents with multi-view coherence. We assigned the value of $m$ to the epoch at around 1000 iterations. For the guidance weight $\omega(t)$, we follow the DreamTime~\cite{huang2023dreamtime} to achieve higher fidelity results. 
Our codes for implementation will be available upon acceptance.

\section{Additional Experiments and Analysis}
\subsection{Additional User Study}
\begin{figure}[!h]
    \centering
    \includegraphics[width=0.95\linewidth]{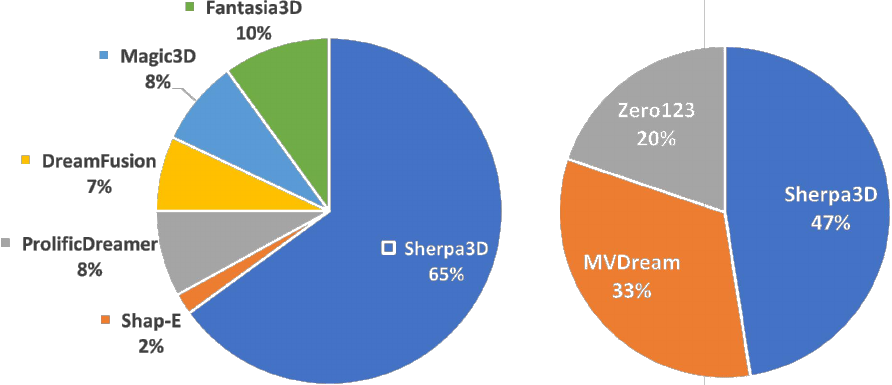}
    \caption{\textbf{User study} of the rate from volunteers' preference for each method in the inset pie chart.}
    \label{fig:user_study}
\end{figure}

To further demonstrate the effectiveness and impressive visualization results of our Sherpa3D, we conducted a more intuitive user study (Figure~\ref{fig:user_study}) on 20 text prompts of five baselines (ShapE~\cite{jun2023shapE}, DreamFusion~\cite{dreamfusion}, Magic3D~\cite{magic3d}, ProlificDreamer~\cite{wang2023prolificdreamer}, Fantasia3D~\cite{chen2023fantasia3d}) and ours. The study engaged 50 volunteers to assess the generated results in 20 rounds. In each round, they were asked to select the 3D model they preferred the most, based on quality, creativity, alignment with text prompts, and consistency. We also compare our method with recent finetuning-based techniques, such as Zero123~\cite{liu2023zero123} and MVDream~\cite{shi2023mvdream}, which utilize more 3D data~\cite{deitke2023objaverse} to retrain a costly 3D aware diffusion model from Stable Diffusion~\cite{stable-diffusion}. We use the same text prompts and settings as mentioned above.
As shown, we observe that Sherpa3D is preferable (65\%) by the raters on average. In other words, our model is preferred over the best of all baselines in most cases. What's more, our Sherpa3D also outperforms than fine-tuning based method in terms of overall performance as they easily suffer from styles (lightning, texture) overfitting~\cite{shi2023mvdream, liu2023zero123}. We believe this is strong proof of the robustness and quality of our proposed method.

\subsection{More Qualitative Results}
\begin{figure}[!h]
    \centering
    \vspace{-1em}
    \includegraphics[width=1\linewidth]{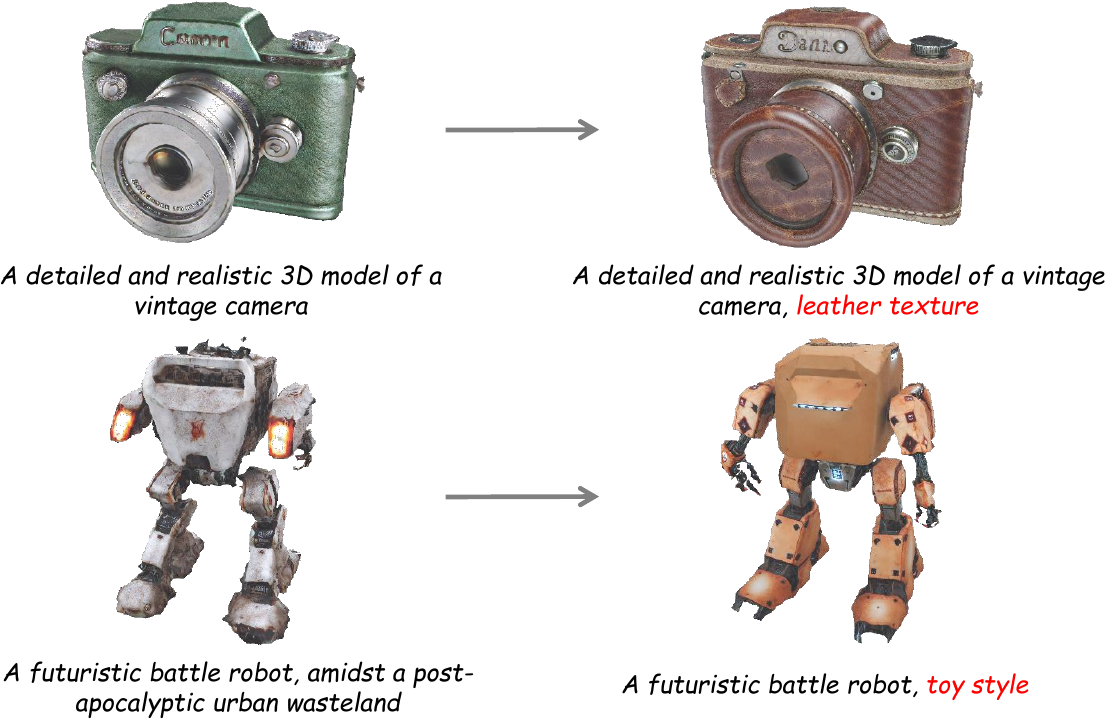}
    \caption{Sherpa3D can be used for flexible editing through a small part of the prompt modification.}
    \label{fig:edit}
    \vspace{-0.8em}
\end{figure}

\begin{figure*}
    \centering
    \includegraphics[width=0.77\linewidth]{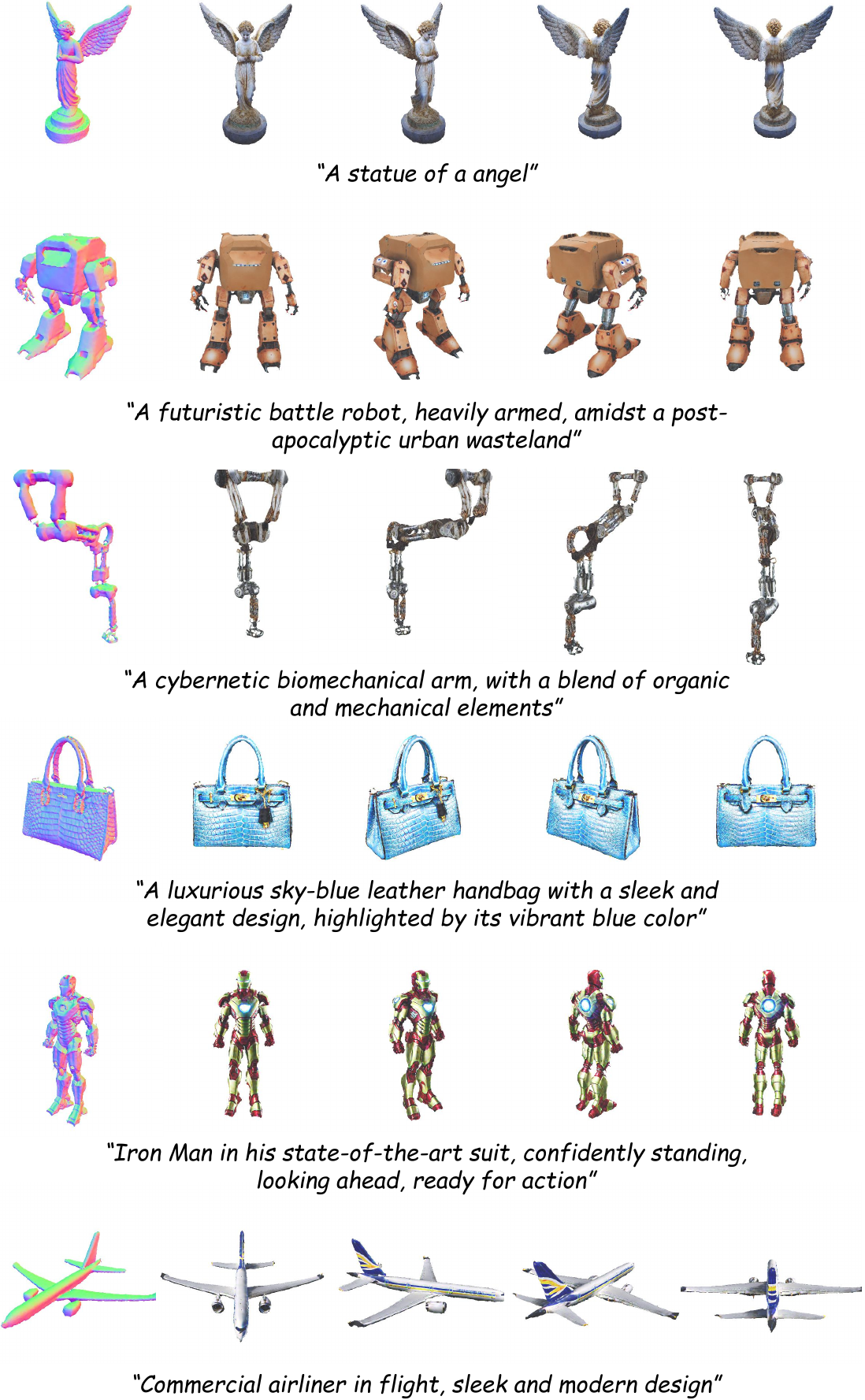}
    \caption{More generated results using our Sherpa3D within 25 minutes. Our work can generate high-fidelity and diversified 3D results from various text prompts, free from the multi-view inconsistency problem.}
    \label{fig:supp_vis_self_1}
\end{figure*}

\begin{figure*}
    \centering
    \includegraphics[width=0.77\linewidth]{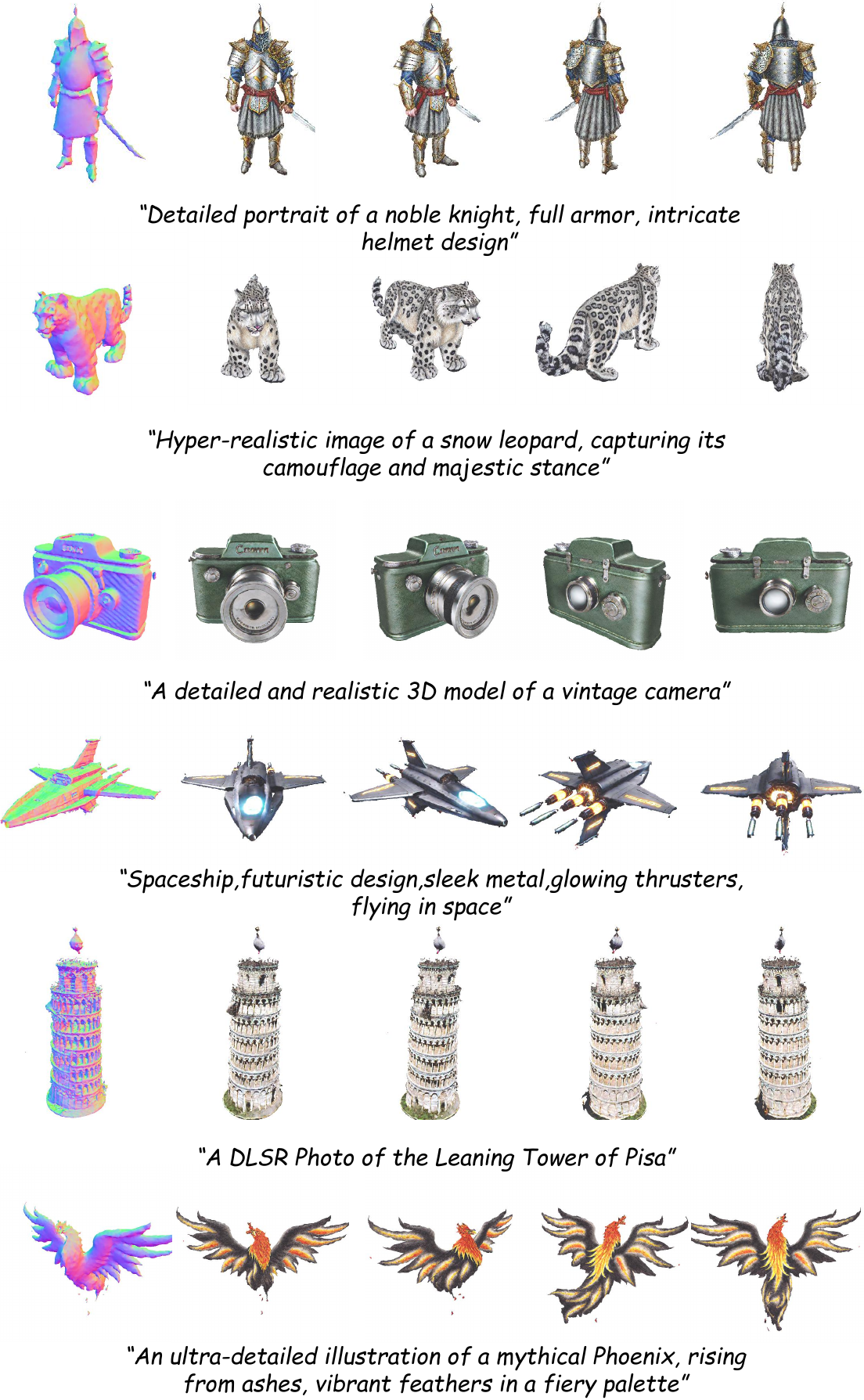}
    \caption{More generated results using our Sherpa3D within 25 minutes. Our work can generate high-fidelity and diversified 3D results from various text prompts, free from the multi-view inconsistency problem.}
    \label{fig:supp_vis_self_2}
\end{figure*}

\begin{figure*}
    \centering
    \includegraphics[width=0.74\linewidth]{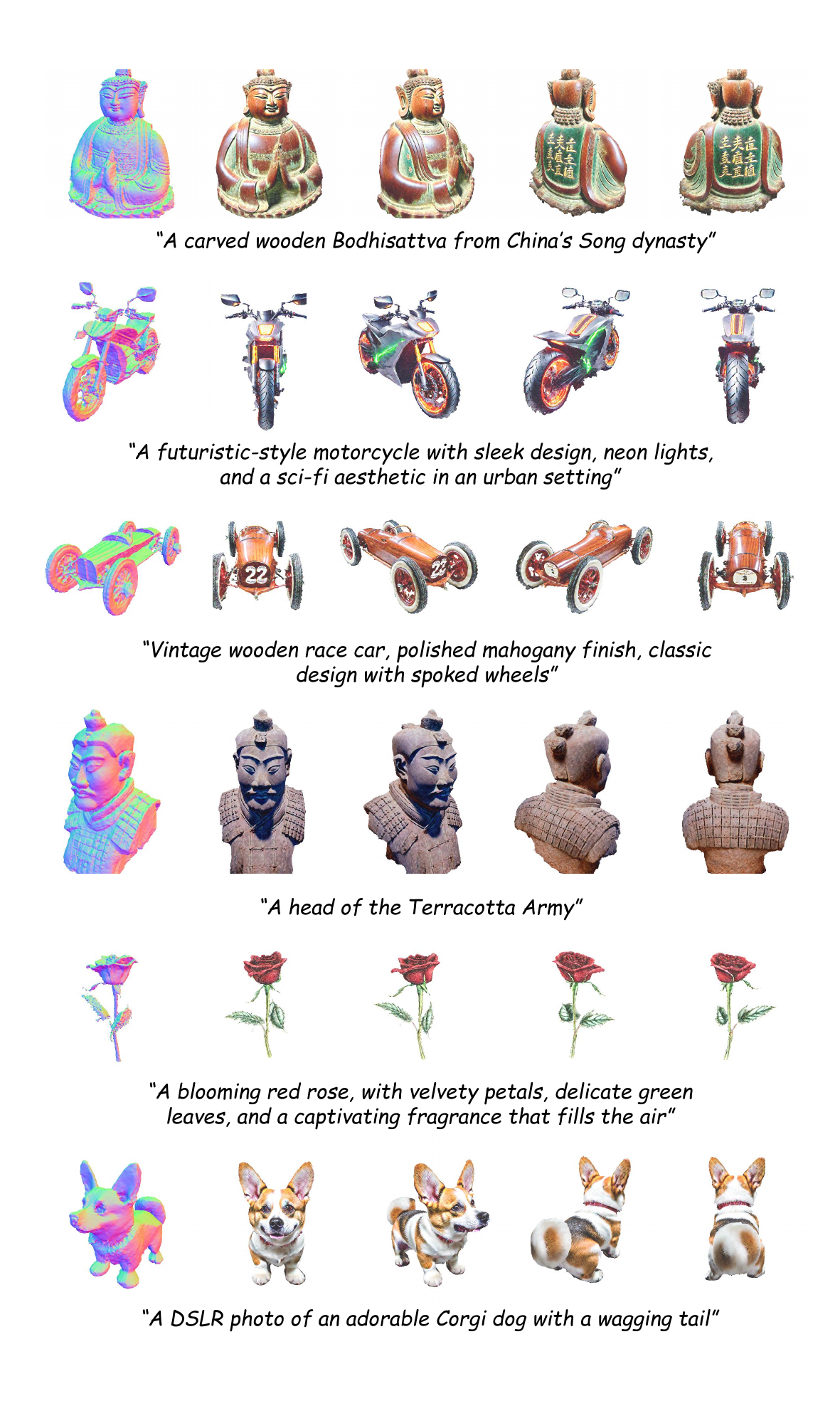}
    \caption{More generated results using our Sherpa3D within 25 minutes. Our work can generate high-fidelity and diversified 3D results from various text prompts, free from the multi-view inconsistency problem.}
    \label{fig:supp_vis_self_3}
\end{figure*}

\noindent \textbf{Sherpa3D. }In Figure~\ref{fig:supp_vis_self_1}, \ref{fig:supp_vis_self_2}, \ref{fig:supp_vis_self_3}, we present more text-to-3D results obtained with Sherpa3D, which can generate high-fidelity, diverse, and 3D-consistent results within 25 minutes.
Besides the impressive 3D consistency and high fidelity, we can also change the style of generated 3D content (Figure~\ref{fig:edit}) by only modifying a small part of the prompt, while preserving the basic structure of 3D content, which is more convenient for users to flexibly edit generated objects.

\begin{figure}[!h]
    \centering
    \includegraphics[width=0.86\linewidth]{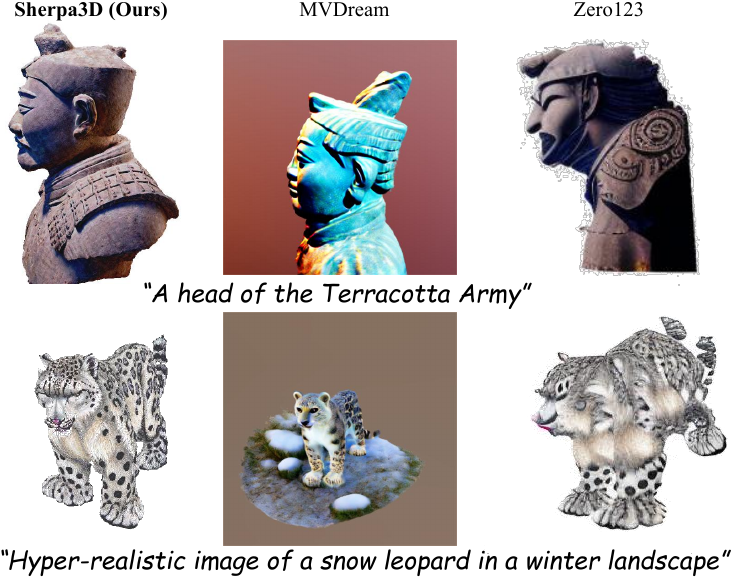}
    \caption{Comparison with MVDream~\cite{shi2023mvdream} and Zero123~\cite{liu2023zero123}. }
    \label{fig:comparison_finetune}
\end{figure}

\begin{figure*}
    \centering
    \includegraphics[width=0.76\linewidth]{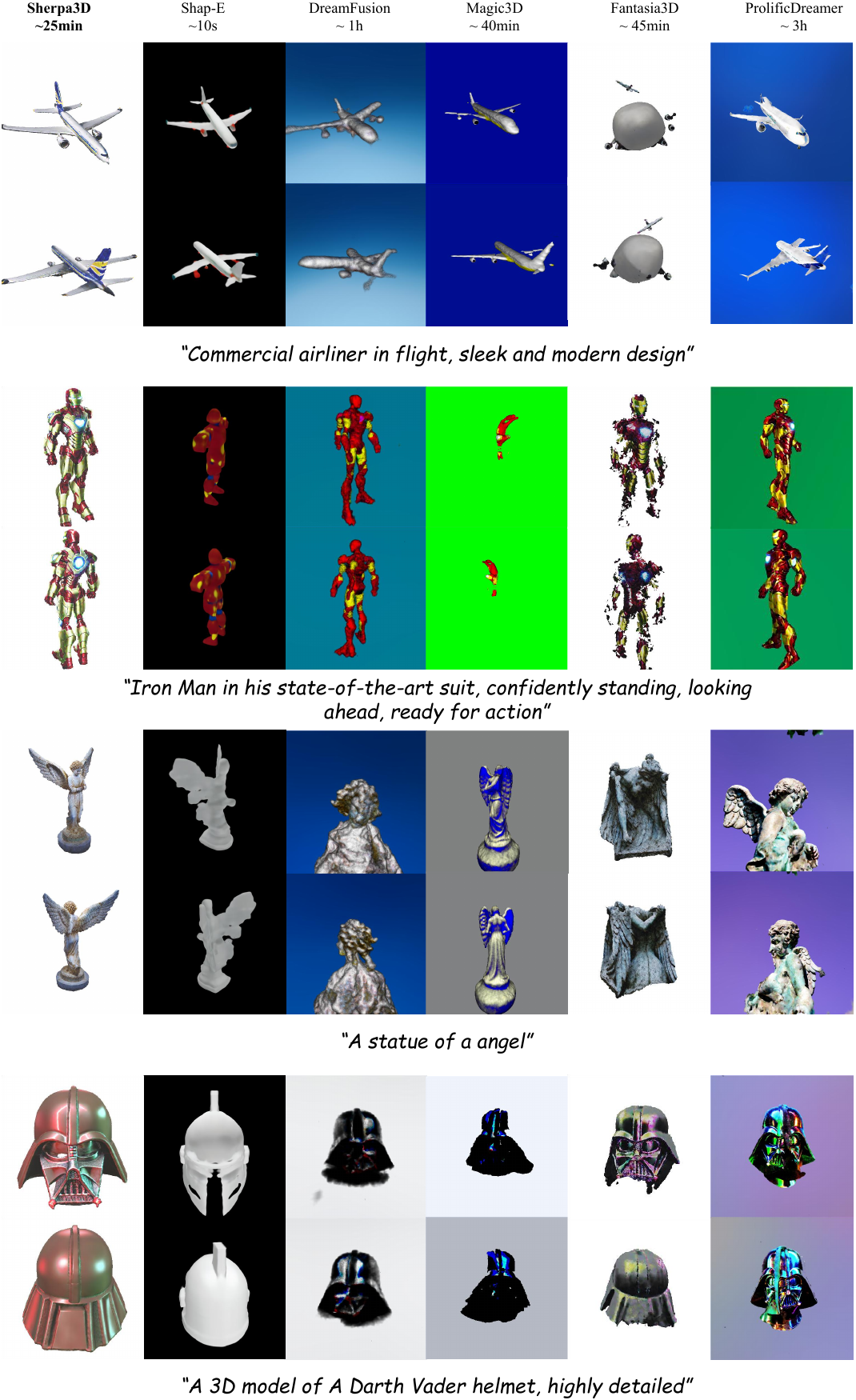}
    \caption{Qualitative comparisons with baseline methods across different views. All methods use \textit{stabilityai/stable-diffsuion-2-1-base} for fair comparison. We observe that baselines suffer from severe multi-face issues while Sherpa3D achieves better quality and 3D coherence.}
    \label{fig:supp_vis_cmp_1}
\end{figure*}

\begin{figure*}
    \centering
    \includegraphics[width=0.75\linewidth]{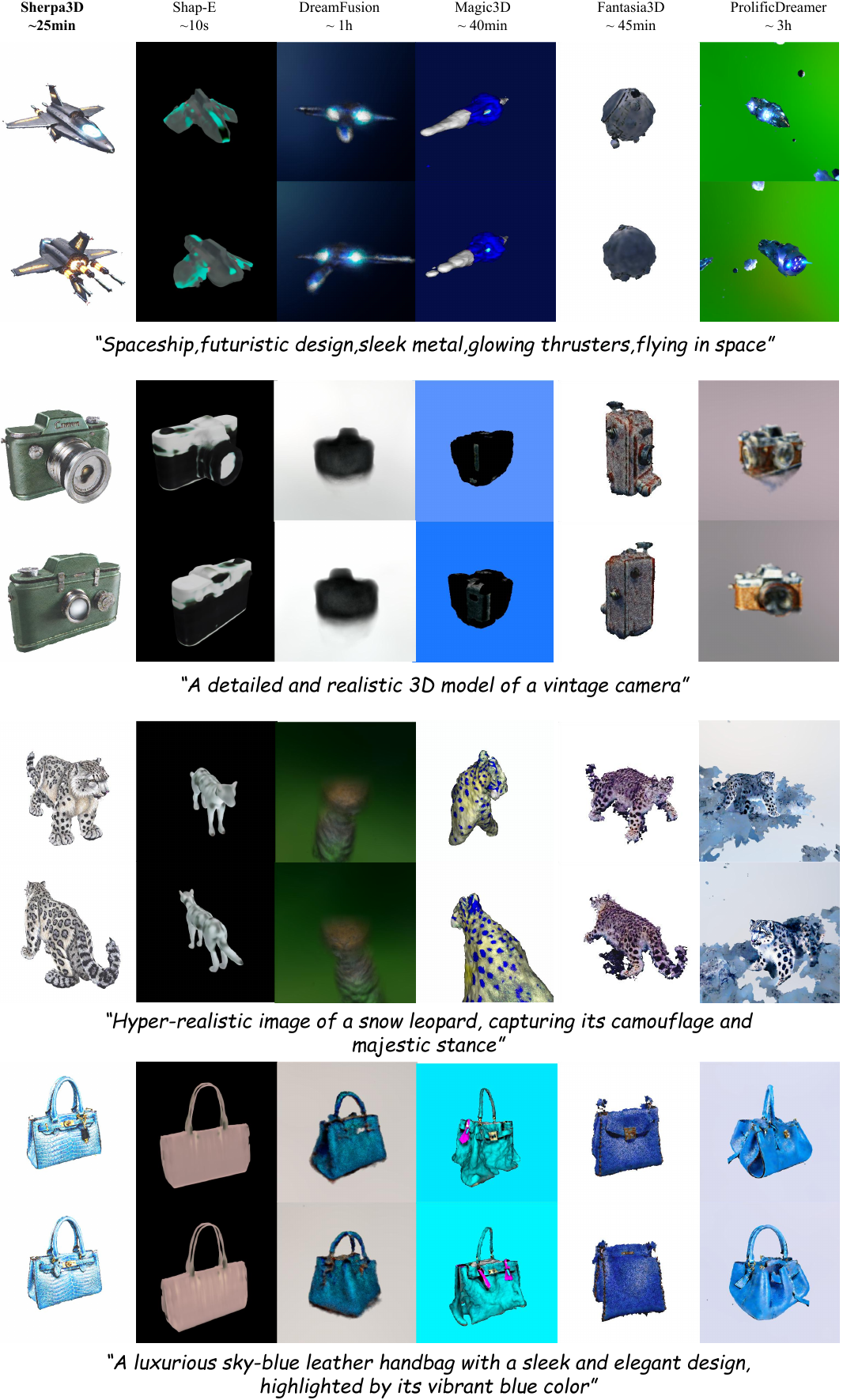}
    \caption{Qualitative comparisons with baseline methods across different views. All methods use \textit{stabilityai/stable-diffsuion-2-1-base} for fair comparison. We observe that baselines suffer from severe multi-face issues while Sherpa3D achieves better quality and 3D coherence.}
    \label{fig:supp_vis_cmp_2}
\end{figure*}

\noindent \textbf{More comparison results.} 
We provide more comparisons with baselines in Figure~\ref{fig:supp_vis_cmp_1}, \ref{fig:supp_vis_cmp_2}.
To further demonstrate the robustness and generalization of our method, we compare our Sherpa3D with Zero123~\cite{liu2023zero123} and MVDream~\cite{shi2023mvdream} in Figure~\ref{fig:comparison_finetune}. Although the concurrent work MVDream and Zero123 can also resolve the multi-view inconsistency issues via fine-tuning a costly viewpoints-aware model, we observe that it is prone to overfit the limited 3D data~\cite{deitke2023objaverse}. Specifically, MVDream generates strange color styles while Zero123 fails in such open-vocabulary prompts.

\end{document}